\documentclass{article}
\usepackage{spconf,amsmath,graphicx}
\usepackage{amssymb}
\usepackage{amsmath}
\usepackage{arydshln}
\usepackage{enumitem}

\setlist{nosep, leftmargin=14pt}
\usepackage{mwe} % to get dummy images
\usepackage{booktabs,multirow}
\usepackage{pifont}
\title{Scribble-Supervised Medical Image Segmentation with Dynamic Teacher Switching and Hierarchical Consistency}
\name{
\parbox{\textwidth}{\centering
Thanh-Huy Nguyen$^{\ast 1}$, 
Hoang-Loc Cao$^{\ast 2}$, 
Dat T. Chung$^{2}$, 
Mai-Anh Vu$^{3}$, \\
Thanh-Minh Nguyen$^{2}$, 
Minh Le$^{2}$, 
Phat K. Huynh$^{2}$, 
Ulas Bagci$^{4\dagger}$
}
\thanks{$\dagger$Corresponding author}
\thanks{$^{\ast}$Equal contribution.}
}

\address{
\parbox{\textwidth}{\centering
$^{1}$Carnegie Mellon University, PA 15213, United States \\
$^{2}$PASSIO Lab, North Carolina A\&T State University, NC 27401, United States \\
$^{3}$University of Houston, TX 77004, United States \\
$^{4}$Northwestern University, IL 60611, United States
}
}

\begin{document}
%\ninept
%
\maketitle
\begin{abstract}
Scribble-supervised methods have emerged to mitigate the prohibitive annotation burden in medical image segmentation. However, the inherent sparsity of these annotations introduces significant ambiguity, which results in noisy pseudo-label propagation and hinders the learning of robust anatomical boundaries. To address this challenge, we propose SDT-Net, a novel dual-teacher, single-student framework designed to maximize supervision quality from these weak signals. Our method features a Dynamic Teacher Switching (DTS) module to adaptively select the most reliable teacher. This selected teacher then guides the student via two synergistic mechanisms: high-confidence pseudo-labels, refined by a Pick Reliable Pixels (PRP) mechanism, and multi-level feature alignment, enforced by a Hierarchical Consistency (HiCo) module. Extensive experiments on the ACDC and MSCMRseg datasets demonstrate that SDT-Net achieves state-of-the-art performance, producing more accurate and anatomically plausible segmentation. 
\end{abstract}
\section{Introduction}\label{sec:intro}

Deep learning has achieved strong success in medical image segmentation, supporting key clinical applications such as diagnosis, treatment planning, and disease monitoring. However, its performance still depends on large-scale, pixel-level annotations, which remain costly and time-consuming for radiologists to produce. To ease this burden, recent works have explored \textit{weakly supervised} paradigms using more affordable labels such as boxes, points, or scribbles. Among these, \textit{scribble annotations}, simple strokes roughly marking target regions, have emerged as a clinically practical alternative.

Despite their practicality, scribble annotations introduce severe supervision sparsity, offering only limited pixel-level information without clear boundary cues. This coarse supervision often leads to \textit{ambiguous pseudo-label propagation}, causing models to misinterpret unlabeled regions and produce inaccurate boundaries, ultimately resulting in overfitting to annotated pixels and poor generalization to unlabeled areas. Prior works have sought to mitigate these issues using \textit{pseudo-labeling} or \textit{consistency regularization} frameworks. For instance, Luo \textit{et al.}~\cite{dmpls} proposed dynamically mixed pseudo-label supervision, Li \textit{et al.}~\cite{scribblevc} introduced vision-class embeddings to enhance semantics, Chen \textit{et al.}~\cite{ail} designed a cross-image matching strategy using a reference set constructed from class-specific tokens and pixel-level features, and Zhang \textit{et al.}~\cite{helpnet} developed a hierarchical perturbation consistency strategy with an entropy-guided pseudo-label ensemble. Related advances in semi-supervised and source-free settings further highlight the importance of \textit{reliable pseudo-label selection} and \textit{uncertainty-aware refinement}~\cite{nguyen2026adaptive,nguyen2026up2d}. Although these methods improved scribble-supervised segmentation, they still face two key issues: (1) \textbf{Pseudo-label noise accumulation}, where erroneous predictions are reinforced during self-training, and (2) \textbf{Lack of adaptive supervision}, as fixed teacher-student frameworks cannot consistently identify the most reliable guidance source.

To overcome these challenges, we propose \textbf{SDT-Net}, a novel scribble-supervised segmentation framework that introduces a \textbf{Dynamic Teacher Switching (DTS)} mechanism and a \textbf{Hierarchical Consistency (HiCo)} module to maximize the utility of sparse supervision. Unlike previous fixed-teacher frameworks, SDT-Net employs two complementary teacher networks and dynamically selects the more reliable one for each iteration based on its performance over the available scribble regions. This adaptive switching minimizes confirmation bias and mitigates pseudo-label noise accumulation. Furthermore, to ensure robust structural learning beyond pixel-level supervision, SDT-Net enforces multi-level feature alignment between the student and the selected teacher through the HiCo module, combining low-level geometric and high-level semantic consistency.

The main contributions can be summarized as follows:
\begin{itemize}
\setlength\itemindent{1.8em}
    \item We propose SDT-Net, a novel dual-teacher framework for scribble-based medical image segmentation, achieving state-of-the-art performance on two public datasets.

    \item We develop a Dynamic Teacher Switching (DTS) module that adaptively selects the most reliable teacher, mitigating noise and confirmation bias from any single teacher.

    \item We introduce a Hierarchical Consistency (HiCo) module that enforces feature-level alignment between the student and the selected teacher at both high- and low-level decoder stages, ensuring the learning of robust structural representations.
\end{itemize}

\begin{figure*}[t]
  \centering
  \includegraphics[width=1.0\linewidth]{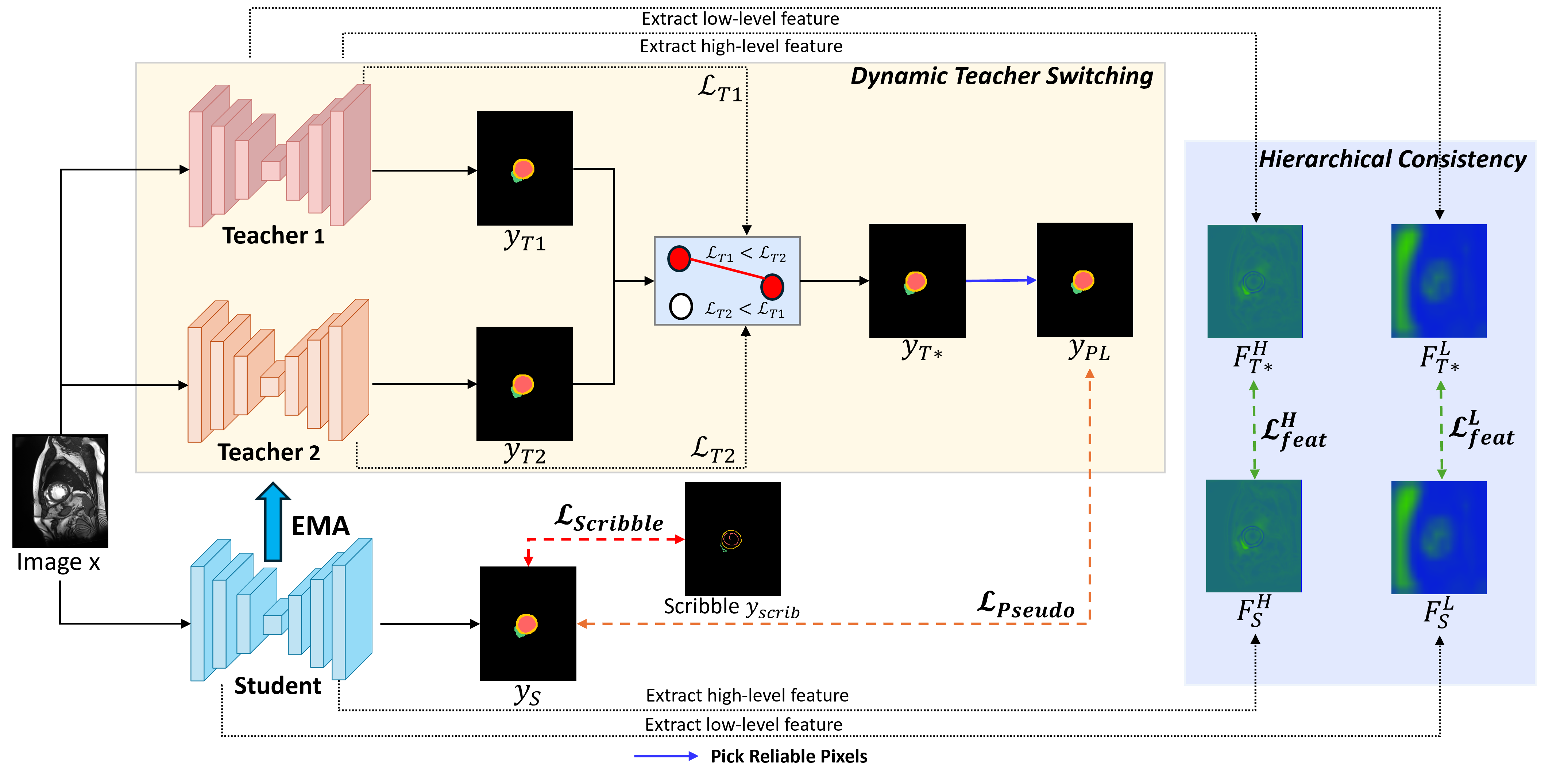}
  \caption{Overview of our SDT-Net. The Dynamic Teacher Switching (DTS) module selects a reliable teacher to generate refined pseudo-labels, while Hierarchical Consistency (HiCo) aligns multi-level features between the student and the selected teacher.}
  \label{fig:framework}
\end{figure*}

\section{METHODOLOGY}
\label{sec:method}
Fig.~\ref{fig:framework} illustrates SDT-Net, which is built upon a single-student, dual-teacher framework. This framework incorporates the Dynamic Teacher Switching (DTS) and Hierarchical Consistency (HiCo) modules. Detailed insights into each module are provided in the subsequent subsections.

% Module scribbe-supervised
\subsection{Scribble-supervised learning}
We train the student network ($y_{S}$) using the ground-truth scribbles ($y_{Scrib}$) via a partial cross-entropy (pCE) loss, denoted as $\mathcal{L}_{Scribble}$ in Fig.~\ref{fig:framework}:
\begin{equation}
\mathcal{L}_{Scribble} = 
\mathcal{L}_{pCE}(y_{S}, y_{Scrib}).
\label{eq:lscribble}
\end{equation}
\noindent $\mathcal{L}_{pCE}$ computes the standard cross-entropy loss only over the set of annotated pixels $\Omega_s$, ignoring all unlabeled regions:
\begin{equation}
\mathcal{L}_{pCE}(y, s) = 
\sum_{i \in \Omega_s} 
\sum_{k \in K} 
 -s_i^k \log y_i^k,
\label{eq:lpce}
\end{equation}
where $i \in \Omega_s$ is an annotated pixel, $K$ is the set of classes, $s_i^k$ is the one-hot ground-truth label, and $y_i^k$ is the network’s predicted probability.

\subsection{Dynamic Teacher Switching}
Our proposed Dynamic Teacher Switching (DTS) module intelligently selects the most reliable teacher (Teacher 1 $T_1$  or Teacher 2 $T_2$ ) on a per-iteration basis, guided by their performance on the sparse scribble data. For each training batch, we evaluate both teachers ($y_{T1}$ and $y_{T2}$) against the ground-truth scribbles $y_{Scrib}$ using the partial CE loss defined in Eq.~\ref{eq:lpce}:
\begin{equation}
\mathcal{L}_{T_1} = \mathcal{L}_{pCE}(y_{T_1}, y_{Scrib})
, \quad 
\mathcal{L}_{T_2} = \mathcal{L}_{pCE}(y_{T_2}, y_{Scrib}).
\label{eq:teacher_loss}
\end{equation}

The teacher with the lower scribble loss is considered more reliable for that specific sample. We designate its softmax probability output as $y_T^*$:
\begin{equation}
y_T^* = 
\begin{cases}
\text{Softmax}(y_{T_1}), & \text{if } \mathcal{L}_{T_1} < \mathcal{L}_{T_2} \\
\text{Softmax}(y_{T_2}), & \text{otherwise}
\end{cases}.
\label{eq:teacher_selection}
\end{equation}

This dynamically selected prediction $y_T^*$ is then used for generating pseudo-label for the student. To obtain high-quality pseudo-label, $y_T^*$ is passed through a Pick Reliable Pixels mechanism, which filters the prediction based on a confidence threshold $\tau$. A pixel $i$ is considered reliable as a pseudo-label only if the teacher's softmax probability vector $(y^*_T)_i$ has a maximum class probability exceed threshold $\tau$. Otherwise, the pixel is ignored. 

% A pixel $i$ is considered reliable only if its maximum prediction confidence exceeds $\tau$:
\begin{equation}
y_{PL} =
\begin{cases}
\arg\max_k (y_T^*)_i, & \text{if } \max_k (y_T^*)_i \ge \tau \\
\text{Ignore}, & \text{otherwise,}
\end{cases}
\label{eq:prp}
\end{equation}
where $k$ is the class index.

This process creates a sparse but reliable pseudo-label map $y_{PL}$, which is used to supervise the student network via a combined pseudo-label loss $\mathcal{L}_{Pseudo}$:
\begin{equation}
\mathcal{L}_{Pseudo} = \frac{1}{2} \Big(
\mathcal{L}_{CE}(y_{S}, y_{PL}) + 
\mathcal{L}_{Dice}(y_{S}, y_{PL})\Big).
\label{eq:pseudo_loss}
\end{equation}
This loss is computed only on the set of reliable pixels identified by Eq.~\ref{eq:prp}.
\subsection{Hierarchical Consistency}
To ensure robust structural representations, our Hierarchical Consistency (HiCo) module enforces feature alignment between the student ($S$) and the reliable teacher ($T^*$) selected by the DTS module (Eq.~\ref{eq:teacher_selection}). This alignment is applied at two distinct decoder stages: low-level features ($F^L$) for local details and high-level features ($F^H$) for semantic context. The consistency loss, $\mathcal{L}_{feat}$, for each level combines an L1-distance loss for geometric similarity with a cosine similarity loss for semantic alignment:

\begin{align}
\mathcal{L}_{feat}(F_S, F_{T^*})
&= \tfrac{1}{2}\Big(\mathcal{L}_{L_1}(F_{S}, F_{T^*}) \notag\\
&\quad + \big[1-\cos(F_{S}, F_{T^*})\big]\Big),
\label{eq:feat_loss}
\end{align}
where $\mathcal{L}_{L_1}$ denotes the L1-distance loss, $F_S$ and $F_{T^*}$ are the corresponding low-level ($F^L$) or high-level ($F^H$) feature maps from the student ($S$) and reliable teacher ($T^*$) decoders, and $\cos(\cdot, \cdot)$ denotes the cosine similarity computed on the flattened feature vectors.

The HiCo loss ($\mathcal{L}_{HiCo}$) is defined as the average of the losses from both hierarchical feature levels:\begin{equation}\mathcal{L}_{HiCo} = \frac{1}{2} \Big( \mathcal{L}_{feat}(F_{S}^{L}, F_{T*}^{L}) + \mathcal{L}_{feat}(F_{S}^{H}, F_{T*}^{H})\Big).\label{eq:hico_loss}\end{equation}

\subsection{Total Loss Function}
The final optimization objective is as follows:
\begin{equation}
\mathcal{L}_{Total} =
\mathcal{L}_{Scribble} +
\mathcal{L}_{Pseudo} +
\mathcal{L}_{HiCo},
\label{eq:total_loss}
\end{equation}
where $\mathcal{L}_{Scribble}$ is the scribble-supervised loss, 
$\mathcal{L}_{Pseudo}$ denotes the pseudo-label supervision loss, 
and $\mathcal{L}_{HiCo}$ represents the hierarchical consistency loss.

While the student network ($S$) is optimized using $\mathcal{L}_{Total}$ via backpropagation, the teacher networks ($T_1$ and $T_2$), 
which share the same architecture as the student, 
do not receive gradients. Instead, only the weights of the reliable teacher ($\theta_{T^*}$), selected by the DTS module (Eq.~\ref{eq:teacher_selection}), are updated at each iteration $t$ via an Exponential Moving Average (EMA) of the student's weights ($\theta_S$) as:
\begin{equation}
\theta_{{T*}}^{t} = \alpha \theta_{{T*}}^{t-1} + (1 - \alpha)\theta_{S}^{t},
\label{eq:ema_update}
\end{equation}\\
where $\alpha$ is the EMA decay rate.

\section{EXPERIMENTS}
\label{sec:exp}

\subsection{Experimental Setup}
\textbf{Datasets}: We evaluated our method on two public benchmarks: ACDC~\cite{acdc} and MSCMRseg~\cite{mscmr}. The ACDC dataset includes 100 2D cine-MRI scans with manual scribble annotations for the right ventricle (RV), left ventricle (LV), and myocardium (MYO), adhering to the standard 70/15/15 train/validation/test split from~\cite{acdcconfig}. The MSCMRseg dataset contains 45 Late Gadolinium Enhancement MR images from cardiomyopathy patients, annotated for the same structures, which we divide into 25/5/15 for training, validation, and testing following~\cite{mscmrconfig}.\\
\textbf{Implementation Details}: We employed UNet~\cite{unet} as the backbone. The framework was implemented in PyTorch 2.2.0 on an Nvidia T4 GPU (16 GB). Input images were resized to $256 \times 256$, and we applied random rotation and flipping for data augmentation. We employed SGD to optimize the student model, with a learning rate of $0.01$, a weight decay of $10^{-4}$ and a momentum of $0.9$. The total iterations, warm-up iterations, batch size, confidence threshold $\tau$, and EMA decay rate $\alpha$ were set $30k$, $12k$, $8$, $0.5$ and $0.999$, respectively. During testing, predictions were generated slice-by-slice, and the student network's output was used as the final result without any post-processing methods. The Dice score was used as the evaluation metric.

% ==========TABLE=========
%  Kết quả trên acdc và mscmr
\begin{table}[htbp]
\centering
\renewcommand{\arraystretch}{1.15}
\resizebox{\columnwidth}{!}{%
\begin{tabular}{l|cccc}
\hline
\textbf{Methods} & \textbf{LV} & \textbf{MYO} & \textbf{RV} & \textbf{Avg} \\
\hline
DMPLS (MICCAI'22)~\cite{dmpls}         & 91.3 & 84.2 & 86.1 & 87.2 \\
ScribbleVC (ACM-MM'23) \cite{scribblevc} & 91.4 & 86.6 & 87.0 & 88.4 \\
ScribFormer (TMI'24) \cite{scribformer} & 92.2 & 87.1 & 87.1 & 88.8 \\
AIL (TIP'25) \cite{ail}         & 92.3 & 85.5 & 86.0 & 87.9 \\
HELPNet (MedIA'25) \cite{helpnet}       & 90.8 & 87.9 & 87.7 & 88.8 \\

\hline
\textbf{Ours}                & \textbf{93.5} & \textbf{89.3} & \textbf{89.6} & \textbf{90.8} \\
\hline
\end{tabular}
}
\caption{Quantitative comparison among scribble-supervised methods on ACDC dataset using Dice score (\%).}
\label{tab:results_acdc}
\end{table}

\begin{table}[htbp]
\centering
\renewcommand{\arraystretch}{1.15}
\resizebox{\columnwidth}{!}{%
\begin{tabular}{l|cccc}
\hline
\textbf{Methods} & \textbf{LV} & \textbf{MYO} & \textbf{RV} & \textbf{Avg} \\
\hline
DMPLS (MICCAI'22)~\cite{dmpls}           & 91.4 & 83.5 & 88.0 & 87.6 \\
ScribbleVC (ACM-MM'23) \cite{scribblevc} & 92.1 & 83.0 & 85.2 & 86.8 \\
ScribFormer (TMI'24) \cite{scribformer} & 89.6 & 80.7 & 81.3 & 83.9 \\
AIL (TIP'25) \cite{ail}         & 90.5 & 80.9 & 84.5 & 85.3 \\
HELPNet (MedIA’25) \cite{helpnet}       & 92.1 & 84.4 & 88.1 & 88.2 \\
\hline
\textbf{Ours}                & \textbf{93.1} & \textbf{85.0} & \textbf{88.8} & \textbf{90.0} \\
\hline
\end{tabular}
}
\caption{Quantitative comparison among scribble-supervised methods on MSCMRseg dataset using Dice score (\%).}
\label{tab:results_mscmr}
\end{table}

\begin{figure}[t]
  \centering
  \includegraphics[width=\linewidth]{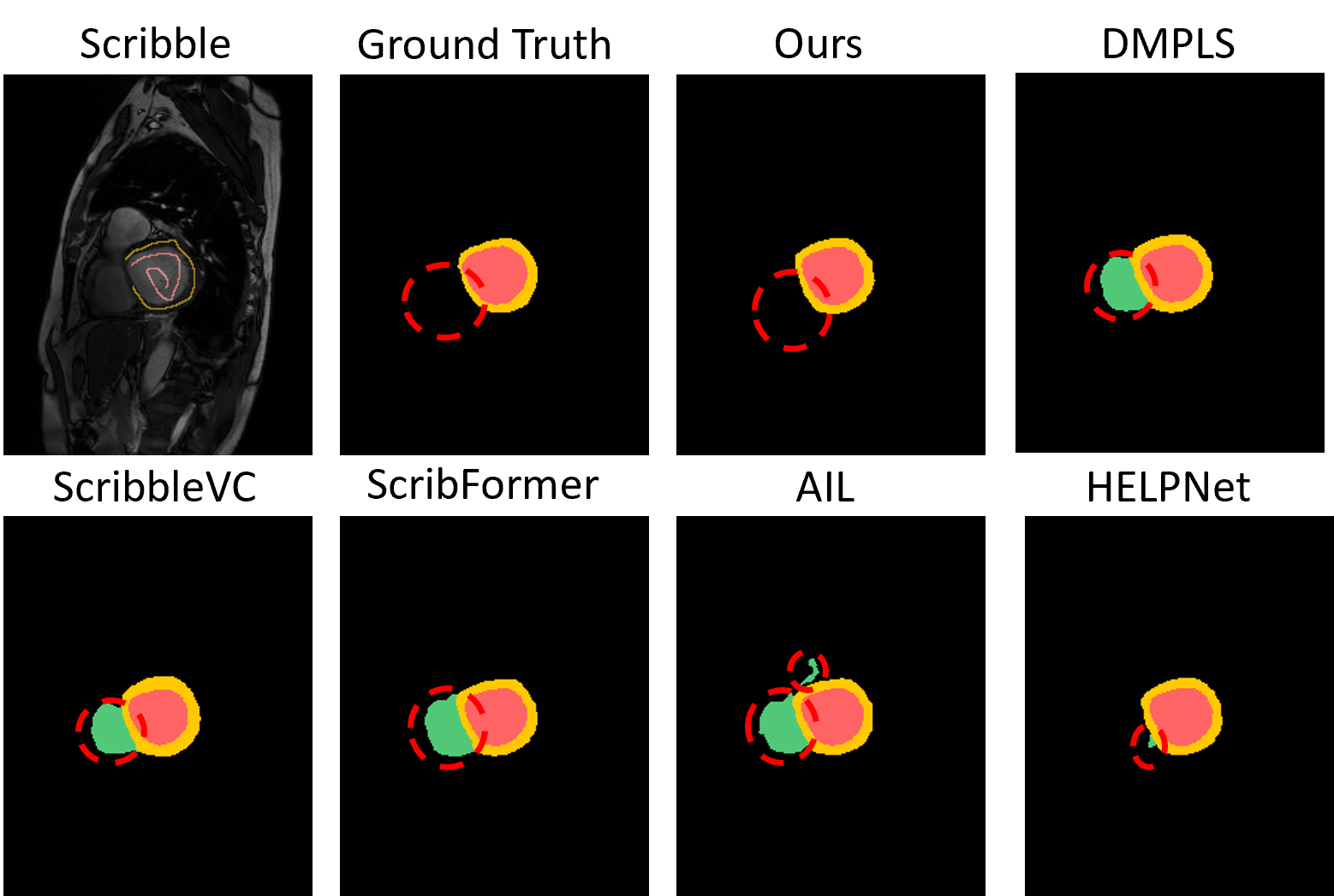}
  \caption{Qualitative results across different scribble-supervised medical image segmentation methods on the ACDC dataset. Red dashed circles highlight regions of different segmentation}
  \label{fig:visual}
\end{figure}
% ABLATION STUDY
\begin{table}[t]
\centering
\setlength{\tabcolsep}{4pt}
\renewcommand{\arraystretch}{1.08}
\resizebox{\columnwidth}{!}{%
\begin{tabular}{l|c|c|c|cccc}
\hline
\textbf{No.~T} & \textbf{PL} & \textbf{PRP} & \textbf{HiCo} & LV & MYO & RV & Avg \\
\hline
\multirow{2}{*}{Single} & \multirow{2}{*}{Teacher} & \ding{55} & \ding{55} & 74.5 & 64.1 & 68.5 & 69.0 \\
                        &                           & $\checkmark$ & $\checkmark$ & 91.8 & 82.7 & 73.6 & 82.7 \\
\hline
\multirow{5}{*}{Dual}   & Avg                        & $\checkmark$ & $\checkmark$ & 92.9 & 84.2 & 86.3 & 87.8 \\
\cline{2-8}
                        & \multirow{4}{*}{DTS}       & \ding{55} & \ding{55} & 92.9 & 83.9 & 87.6 & 88.1 \\ 
                        &                            & $\checkmark$ & \ding{55} & 91.3 & 83.3 & 81.3 & 85.3 \\                        
                        &                            & \ding{55} & $\checkmark$ & 92.4 & 84.5 & 88.7 & 88.5 \\
                        &                            & $\checkmark$ & $\checkmark$ & \textbf{93.1} & \textbf{85.0} & \textbf{88.8} & \textbf{90.0} \\
\hline
\end{tabular}%
}
\caption{Effectiveness evaluation of proposed components on MSCMRseg dataset. No.~T: number of teachers; PL: pseudo-labeling strategy (Avg: average, DTS: dynamic teacher switching); PRP: pick reliable pixels; HiCo: hierarchical consistency module.}
\label{tab:ablation}
\end{table}

% =====END TABLE======

\subsection{Experimental Results and Discussion}
Table~\ref{tab:results_acdc} and Table~\ref{tab:results_mscmr} present a comparative analysis of our SDT-Net's performance against several scribble-supervised methods on the ACDC and MSCMRseg datasets. Our method achieves the highest average Dice scores of 90.8\% on ACDC and 90.0\% on MSCMRseg, outperforming the previous best method (HELPNet) by 2.0\% and 1.8\%, respectively. Furthermore, our method surpasses all compared methods. On the ACDC dataset, it demonstrates significant improvements of 3.6\% (vs. DMPLS), 2.4\% (vs. ScribbleVC), 2.0\% (vs. ScribFormer), 2.9\% (vs. AIL), and 2.0\% (vs. HELPNet). Similar superiority is observed on the MSCMRseg dataset, where our method outperforms AIL and HELPNet by 4.7\% and 1.8\%, respectively.

The qualitative comparison in Fig.~\ref{fig:visual} highlights that while compared methods consistently suffer from over-segmentation in the right ventricle (RV) region, our SDT-Net generates the most accurate segmentation that adheres to the ground truth boundaries and preserves anatomical topology.

We also investigate the impact of our proposed components on the MSCMRseg dataset. As illustrated in Table~\ref{tab:ablation}, our full model, which integrates a dual-teacher framework with DTS, PRP, and HiCo, achieves the highest average Dice score of 90.0\%. This demonstrates a significant 7.3\% improvement over the single-teacher baseline (82.7\%), confirming the effectiveness of our dual-teacher design. Isolating the pseudo-labeling strategy, our proposed DTS (88.5\%) provides a clear advantage over a simple averaging (Avg) policy (87.8\%). Further analysis presents a notable finding: relative to the 88.5\% DTS baseline, adding either PRP (85.3\%) or HiCo individually fails to improve segmentation performance. This trend proves that both PRP and HiCo are necessary to effectively leverage the sparse annotations.

\section{CONCLUSION}
\label{sec:typestyle}
This paper introduces SDT-Net, a novel dual-teacher framework for scribble-supervised medical image segmentation. Our method leverages a Dynamic Teacher Switching (DTS) module to select the most reliable teacher, which then guides the student through both high-confidence pseudo-labels and a Hierarchical Consistency (HiCo) constraint. Extensive experiments on the public ACDC and MSCMRseg datasets validate the framework's state-of-the-art performance, producing more robust and topologically accurate segmentations.

\section{ACKNOWLEDGMENT}
This project is partially supported by NIH R01-HL171376 and NIH U01-CA268808.

% \section{ACKNOWLEDGMENTS}
% \label{sec:acknowledgments}
% Thank you

% References should be produced using the bibtex program from suitable
% BiBTeX files (here: strings, refs, manuals). The IEEEbib.bst bibliography
% style file from IEEE produces unsorted bibliography list.
% ------------------------------------------------------------------------- 
\bibliographystyle{IEEEbib}
\bibliography{strings,refs}

\end{document}